# Automated Workers' Ergonomic Risk Assessment in Manual Material Handling using sEMG Wearable Sensors and Machine Learning

**Srimantha E. Mudiyanselage [1], Phuong H.D. Nguyen [2], Mohammad Sadra Rajabi [3] and Reza Akhavian [3,\*]**

[1] School of Engineering, California State University East Bay, 25800 Carlos Bee Blvd, Hayward, CA 94542, United States; sedirisinghemudiyanselage@horizon.csueastbay.edu

[2] Hole School of Construction Engineering, Department of Civil and Environmental Engineering, University of Alberta, 9211 116 St NW, Edmonton, AB T6G 1H9, Canada; phnguye1@ualberta.ca@ualberta.ca

[3] Department of Civil, Construction, and Environmental Engineering, San Diego State University, 5500 Campanile Dr, San Diego, CA 92182, United States; mrajabi1431@sdsu.edu; rakhavian@sdsu.edu

\* Correspondence

**Abstract:** Manual material handling tasks have the potential to be highly unsafe from an ergonomic viewpoint. Safety inspections to monitor body postures can help mitigate ergonomic risks of material handling. However, the real effect of awkward muscle movements, strains, and excessive forces that may result in an injury may not be identified by external cues. This paper evaluates the ability of surface electromyogram (EMG)-based systems together with machine learning algorithms to automatically detect body movements that may harm muscles in material handling. The analysis utilized a lifting equation developed by the U.S. National Institute for Occupational Safety and Health (NIOSH). This equation determines a Recommended Weight Limit, which suggests the maximum acceptable weight that a healthy worker can lift and carry as well as a Lifting Index value to assess the risk extent. Four different machine learning models, namely Decision Tree, Support Vector Machine, K-Nearest Neighbor, and Random Forest are developed to classify the risk assessments calculated based on the NIOSH lifting equation. The sensitivity of the models to various parameters is also evaluated to find the best performance using each algorithm. Results indicate that Decision Tree models have the potential to predict the risk level with close to 99.35% accuracy.

## Introduction

While the introduction of automation and robotics alleviated the frequency and intensity of manual work, many industries such as construction still rely on labor manual work as the main source of production. Work-related musculoskeletal disorders (WMSDs) are among the most reported jobsite injuries [1,2]. Labor manual activities at construction sites offer a high risk of exposure to vibration, repetitive movements, forceful exertions, non-neutral trunk postures, and awkward motions for extended durations which increase the risk of WMSDs. Inability to prevent such insecure postures and movements can translate to escalated risk of injuries while deteriorating workers' physical health, morale, and the quality of life as well as the employer's safety records and productivity. Prevention is undoubtedly more desirable than cure and ergonomic risks can be successfully eliminated by raising workers' self-awareness of hazardous postures and body movements.

The advancement of technology in miniaturized electronics and powerful cloud-based software has enabled successful applications of sensor-based Internet-of-Thing (IoT) platforms in construction jobsites [3,4]. Some of the recent applications of such IoT systems enabled quantifying workers body motions using wearable sensors. Many of those past studies used Inertial Measurement Units (IMUs) and vision-based systems to examine construction resources' activities [5]. While those data collection techniques proved successful to detect workers movements, certain risks cannot be identified unless direct measurement of muscle activities are performed. This is because while an activity may not seem harmful, the real effect of awkward muscle movements, strains, and forces may result in hazardous ergonomic situations that cannot be identified using IMUs or vision-based systems. This paper presents the results of a series of empirical analyses that leverage surface electromyography (sEMG) sensors to directly measure the human muscle electrical impulses during activities that involve lifting objects. The hypothesis of the research is that sEMG can characterize ergonomically hazardous body postures and muscle extremes. To evaluate the hypothesis, the experiments results have been evaluated in the context of the National Institute for Occupational Safety and Health (NIOSH) Lifting Equation [6].

## 2. Literature Review

Understanding the nature and preventing exacerbation of musculoskeletal disorders (MSDs) as a result of work-related injuries have been subject to many past studies. MSDs are referred to as the situations involving the nerves, tendons, muscles, and supporting structures of the body are injured. In particular, low back pain, shoulder injuries, and distal upper extremity disorders, including tendonitis, epicondylitis, and carpal tunnel syndrome, can be named as such disorders [7]. Work-related MSDs or WMSDs, are among the most reported jobsite injuries that can negatively affect workers' health, well-being, and productivity [1,2,8]. This has been identified by many previous past studies conducted in collaboration with national and international safety organization. For example, in a research study in 2004, Waters [7] summarized the findings of two research agendas to increase our understanding of the approaches that can prevent this kind of disorders. The first agenda was based on the data gathered from several hundred practitioners and safety and health specialists representing industry, labor, and academia, and developed by the National Institute for Occupational Safety and Health's (NIOSH) National Occupational Research Agenda (NORA) MSD team. In the second agenda, which was developed by the National Research Council (NRC) and the Institute of Medicine's (IOM) National Panel on MSDs and the Workplace, data had been collected from leading researchers in the fields of medicine, information science, and ergonomics. The results clearly showed that work-related injuries and illnesses represent a considerable health problem for the U.S. industrial labor force [7]. In another study by Pascual and Naqvi [9] the ergonomics risk assessment methods used by Joint Health and Safety Committees (JHSCs) have been investigated. It is concluded that most JHSC curricula have minimal ergonomics scope; thus, JHSCs rely mostly on injury reports and worker complaints to assess ergonomics risk, and indeed most ergonomics analysis tools available require some ergonomics knowledge.

There are research studies on proactive monitoring of industrial workers' awkward postures and non-ergonomic motions that date back to 1990s. Schoenmarklin et al. [10] monitored the acceleration in the flexion/extension as the best kinematic parameter for evaluation of the low and high incident rates of hand/wrist Cumulative Trauma Disorders

(CTDs). However, recent studies focused more on using automated methodologies. Such approaches often take advantage of the advancements in wearable sensors or vision-based monitoring methods to detect such hazardous situations [11,12]. Recently, wearable sensing technologies have provided opportunities to gather near real-time data to analyze workers' safety and health situations [13]. These approaches are often characterized by features such as being low-cost, easy to use, highly accuracy, and non-intrusive [14]. In a study aiming at reviewing MSDs in the construction industry, Valero et al. [15] indicated that the subjectivity and lack of accuracy of visual assessment demand replacing such observations with more accurate and precise posture measurement devices and methods. In some recent studies, in order to identify jobsite workers activity, accelerometers embedded in smart mobile phones have been used [12–14,16–19]. An artificial neural network (ANN)-based models have been developed for identifying falls and manual material handling activities with a high accuracy using the smartphone installed on workers' bodies by Akhavian and Behzadan [16]. Yang et al. [19] presented a method based on smartphone sensor data acquisition and the concept of labor intensity to evaluate construction workers' workloads. A sensor application based on the smartphone platform was created to effectively measure labor intense activities so that the application could track construction workers' movement data unobtrusively.

Recently, IMU sensor-based activity identification models have been explored for a diverse set of applications in the construction industry such as work sampling and fall detection to address practical implementation issues with smartphones [20–23]. Furthermore, the IMU sensor-based models have been vastly utilized towards enhancing ergonomic and safety aspects of construction activities [12,16,24,25]. Yan et al. [25] designed a motion warning system a real-time motion to detect predefined thresholds of hazardous ergonomic postures and alert the worker's smartphones. Even though this IMU system recognized movement directions, angles, and rotation it failed to recognize the real muscle stress and power. In order to validate the system, robust clinical motion data output and sufficient alarm ringing were utilized in both laboratory and field experiments on a construction site located in Hong Kong. The results show that the proposed system assists construction workers to prevent WMSDs without disturbance and interruption during the operations [25]. Jahanbanifar and Akhavian [24] modeled body movements and physical activities, including pushing and pulling, in laboratory-scale experiments. The performed force was measured by a work simulator tool, and a smartphone sensor installed on the active arm collected accelerometer data. An ANN was trained with the accelerometer data and the force levels. The testing data results reveal that the trained model can predict the force level with over 87.5% accuracy [24]. Generally, such methods use IMUs as a wearable tool that detects acceleration produced due to performing a specific activity. The produced data are used to train machine learning algorithms that enable detecting the identified hazardous activities in examples not used during the training of the model. Such models are then evaluated for their generalization, and their performance is enhanced through the use of more training examples and sophisticated prediction algorithms.

Despite the fact that sensors provide detailed information, not all sensors can be utilized in the construction industry due to the dynamic nature of construction activities [26]. Based on previous research, the ideal sensors for construction applications should have some unique characteristics such as being simple and easy to wear, unobtrusive, affordable, and wireless. Moreover, the sensor should provide valid data and import minimal or no preprocessing for noise cancellation. Thus, identifying a reliable sensor for classifying and monitoring construction activities is crucial, and it can help develop health monitoring systems to prevent WMSDs. In recent studies, attaching sensors using armbands and wristbands have been identified as an affordable, non-invasive, lightweight, and wireless wearable data collection method that is suitable to gather workers' forearm sEMG and inertial measurement unit (IMU) data [26].

Many researchers have employed these data collection methods for different applications in various fields; nevertheless, the use of sEMG sensors has not been fully investigated for the ergonomic risk evaluation in the context of construction workers' safety and health. Neck disorders have been the subject of a study where sternocleidomastoid and the upper trapezius muscle activities were monitored using sEMG sensors [27]. The experiments, however, were intrusive, and the data collection probes were not wearable. In another study, sEMG systems were operated to examine the lower back movements in prefabricated panels erection [8]. The predictive ability of the developed model was somewhat limited due to limitations, such as using only one motion segment. Matsumura et al. experimented with a wrist-band type sEMG sensor to analyze and recognize sEMG data using a Neural Networks (NN) model. This research was limited only to identify a few wrist movements and did not address the construction industries' safety requirement [28]. In a recent study, researchers have assessed construction workers' fatigue level using sEMG systems [29]. Results has shown a considerable difference in sEMG parameters while subjects represented different fatigue levels. Results also proved the workability of the wearable sEMG to evaluate workers' muscle fatigue as a means for assessing their physical stress on construction sites [29]. Bangaru et al. [14] evaluated the data quality and reliability of forearm sEMG and IMU data from a wearable sensor for jobsite activity classification. In order to achieve this goal, seven experiments have been conducted. The experiments' results show that the arm-band sensor data quality is close to the conventional sEMG and IMU sensors with perfect relative and absolute reliability between trials for all the considered activities.

Despite the potential of wearable sEMG systems to identify different activities risk, the feasibility of a wearable sEMG system to identify heavy weightlifting activities has received very little attention. To address this issue, and in order to identify high-risk lifting activities in jobsites, the present study has is proposed where an sEMG-based system is used in a total of 54 experiments to collect data from closely monitored lifting activities. The following sections delineates the developed methodology.

## 3. Research Methodology

In the presented study, a series of experiments were conducted in a controlled environment to measure the muscle activity while performing manual material handling. An sEMG system was used to detect muscle movements in static lifting tasks, which frequently occur as part of material handling. The risk assessment was conducted using the NIOSH lifting equation. In this section, first, a brief overview of the EMG system used and the NIOSH lifting equation are provided, and then the experiments are described in detail.

### 3.1. EMG System

EMG sensors detect micro-electrical signals generated by muscles once muscle-cells are neurologically activated by brain commands. The superior data quality made EMG systems increasingly popular in industries where significant human movement is expected. Most importantly, its potential to measure internal muscle movements and strains as well as to identify extremes make it an ideal tool to assess postural hazards. EMG sensors are categorized into two types and the first type is known as 'insertion-electrodes'. By inserting a needle-like-electrode into a deep, targeted muscle, precise muscle EMG movements can be measured. Since this procedure always requires a medical expert for the insertion and it has some risk of bleeding, infections, and nerve injuries, it is not suitable for everyday use by workers.

Therefore, the less expensive and safer second type, 'Surface-type-EMG' or sEMG sensors with more accurate 'wet-type' electrodes were selected for the experiment. When a muscle requires more force, additional motor units will be added, and then sEMG signal becomes denser visually with increased amplitude. A weighted sum of the action potential is therefore recorded as the sEMG value since the muscle fibers are located deep and far from the surface.

The proposed system consists of an input stage, a signal processing stage, and an evaluation stage. The real-time micro-electrical signals from sEMG electrodes are transferred to the receiver via Bluetooth and micro-voltages of data will be amplified and converted to Digital from Analog. The software algorithm processes captured muscle motion data and translated muscle movements into meaningful parameters in real-time. The Noraxon Mini DTS system with 2 wireless sEMG muscle sensors was used for this experiment along with Myo-Muscle software version MR3 3.14.28. A webcam was also used to record tests for later synchronization of the collected data for analysis.

*3.2. NIOSH Lifting Equation*

The top five ergonomic issues in industries with heavy manual work include working over a long period of time at the same position, awkward twisting or bending, working at cramped positions, handling heavy tools, equipment, or materials in an unsafe manner, and working with an injury. Major hazardous ergonomic postures, longer holding times, and repetitive work in a similar posture for an extended duration can clearly increase the risks of future WMSD. All such instances are reflected in an equation devised by NIOSH to prescribe the allowable limit for lifting weights. The NIOSH lifting equation entails a series of variables and multipliers to consider the relative position of a person to the object they lift as well as the quality of the workers grip, frequency and duration of lifting activity, and twisting requirement [6]. Using NIOSH lifting equation, the Recommended Weight Limit (RWL) which prescribes the maximum acceptable weight that a healthy employee could lift can be determined (see Equation 1).

$$RWL = LC\ (51) \times HM \times VM \times DM \times AM \times FM \times CM \qquad (1)$$

where LC is a load constant equal to 51 lbs. and the rest of the parameters are multipliers (M) including H as the horizontal location of the object relative to the body, V as the vertical location of the object relative to the floor, D as the distance the object is moved vertically, A as the asymmetry angle or twisting requirement, F as the frequency and duration of lifting activity, and C as the coupling or quality of the workers grip on the object. Table 1 and Figure 1 adopted from [6] describes the variables used in this equation.

**Table 1.** NIOSH Lifting Equation Multipliers

| Unit | Load Constant (LC) | Horizontal Multiplier (HM) | Vertical Multiplier (VM) | Distance Multiplier (DM) | Asymmetric Multiplier (AM) | Frequency Multiplier (FM) | Coupling Multiplier (CM) |
|---|---|---|---|---|---|---|---|
| Metric | 23 kg | (25/H) | 1-(.003\|V-75\|) | 0.82+(4.5/D) | 1-(.0032A) | [1] | [2] |
| U.S. Customary | 51 lb | (10/H) | 1-(.0075\|V-30\|) | 0.82+(1.8/D) | 1-(.0032A) | [1] | [2] |

[1] determined based on the frequency of lifts and the vertical distance (V).

[2] determined based on the coupling quality and the vertical distance (V).

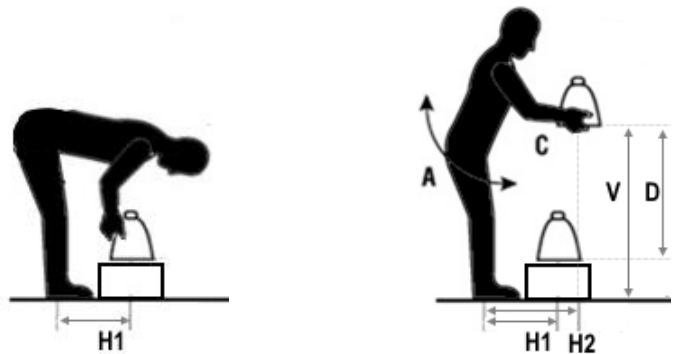

**Figure 1.** Description of the variables used to calculate NIOSH lifting equation multipliers.

A Lifting Index (LI) value is also calculated by dividing the actual weight to be lifted by the RWL (see Equation 2).

$$LI = Weight \div RWL \qquad (2)$$

Therefore, an LI value below 1.0 does not pose a significant risk to a healthy employee. A value exceeding 1.0 means an increased risk with the task and an LI value above 3.0 states that the task as a high-risk to a healthy worker. As the LI increases, the level of upper back injury risk and associated MSD risk increases correspondingly.

*3.3. Experiments*

In order to determine the most sensitive muscles to the lifting tasks, simple and straight body movements related to hands, forearms, arms, legs, thighs, and back were performed to evaluate the change in the signal received from sEMG sensors connected to those muscles. A member of the research team performed a few pre-defined safe lifting activities according to NIOSH lifting requirements. Sensors connected to muscles other than the lower back and upper back produced false positives, false negative, and nondiscriminatory and inconsistent results. As such, the upper back (i.e., Thoracic or TH) muscles were chosen for data collection.

All the lifting tasks were performed by a member of the research team, a male at age 40, height 5'8", and weight 180 lbs. lifting multiple weights. A dumbbell set with adjustable weights from 10 lbs. to 35 lbs. and a size of approximately 14" ×9" ×9" was used as the 'Weight'. During the entire experiment, the lifting speed and frequency were kept constant with some pre-training to minimize the impact. The total number of 54 tests have been done using weights of 10 lbs., 15 lbs., 20 lbs., 30 lbs., and 35 lbs. and with H values equal to 15 and 17. Each test was repeated 10 times at 10 lifts per minute except for tests with 35 lbs. and H17 that was done only 5 times to limit the potential of any injury. In addition, one experiment with the 10 lbs. weight was removed due to incomplete data collection. Repeating each weight lift several times was done to collect more data with higher generalizability potential of the machine leering models to be developed. The overall experiment was completed within 1 hour including the rest breaks. Table 2 describes the lifting experiment variables. All the variables except for the H in the last set of experiments were kept constant. A slightly higher H value of 17" was used in the last set of experiments.

**Table 2.** NIOSH Lifting Equation Multipliers According to Table 1 and Figure 1

|  | H | V | D | A | C | F | Duration |
|---|---|---|---|---|---|---|---|
| **Origin – Lifting the weight** | 15" | 14" | 18" | 0 | Good | 10/min | 1hr/day |
|  | HM | VM | DM | AM | CM | FM |  |
|  | 0.67 | 0.88 | 0.92 | 1 | 1 | 0.45 |  |
|  | H | V | D | A | C | F | Duration |
| **Destination – Placing the weight** | 24" | 32" | 18" | 0 | Good | 10/min | 1hr/day |
|  | HM | VM | DM | AM | CM | FM |  |
|  | 0.42 | 0.99 | 0.92 | 1 | 1 | 0.45 |  |

According to the Equation 1 and based on the multipliers provided in Table 2, the RWL is calculated as 12.4 for the first five sets of experiments. Therefore, the values of LI can be calculated from Equation 2 and are shown for each experiment in Table 3.

**Table 3.** Variable Weights used in the Experiments to Calculate the RWL and LI

| Experiments | Load (LBS) | H | RWL | LI | Risk |
|---|---|---|---|---|---|
| 1-9 | 10 | 15 | 12.40 | 0.8 | Nominal Risk |
| 10-19 | 15 | 15 | 12.40 | 1.2 | Nominal Risk |
| 20-29 | 20 | 15 | 12.40 | 1.6 | Increased Risk |
| 30-39 | 30 | 15 | 12.40 | 2.4 | Increased Risk |
| 40-49 | 35 | 15 | 12.40 | 2.8 | High Risk |
| 50-54 | 35 | 17 | 11.00 | 3.2 | High Risk |

As shown in Table 3, a total of 54 experiments were conducted to collect sEMG data from closely monitored lifting activities. Using the NIOSH lifting equation, the experiments were placed under three risk categories using Lis from 0.8 to 3.2. First, 19 tests were designed to have 'Nominal Risk' with an LI value below or equal to 1.2. The next 20 tests were designed as 'Increased Risk' activities by increasing the LI value from 1.2 to 2.4. The last 15 activities were 'High Risk' lifting activities with an LI value above 2.8.

## 4. Data Analysis

Experiments 1 to 9 were used as a proof-of-concept to observe potential changes in the received signal from the upper back sensor as a result of weightlifting activities. The lifter started by lifting a dumbbell-set weighted 10 lbs. located on a bench and placed it on a higher level. Since only the magnitude of the received signal is sufficient to evaluate the effect on muscles, absolute values of the collected sEMG data are used. Averaged peak absolute value of the signals received was calculated based on 150µV for the first 9 experiments. Also, fast Fourier transform (FFT) was used to observe the effect on the signal frequency domain in addition to the raw data in time domain. This resulted in 22.65µV in the frequency-domain space for the average peak value. Similar to the first 9 experiments, the following five sets of experiments were conducted with a fixed weight of 15, 20, 25, 30, and 35 lbs.

*4.1. Machine Learning Classification Algorithms*

In order to establish a model to classify ergonomic risk based on the lifting activities, the performance of four machine learning classification algorithms, including Decision Tree, Support Vector Machine (SVM), K-Nearest Neighbor (KNN), and Random Forest have been examined. In this study, *weight*, and H as well as the *maximum* value, *minimum*, *average*, *median* of modified FFT, and *standard deviation* for each test have been used as the features for training the machine learning classification models. The study was also designed to examine the sensitivity of the models to various window sizes for feature extraction. In particular, three classification scenarios were conducted with 1-second window sizes segmentation (total data points n = 425), 0.5-second window sizes segmentation (n = 845), and 0.25-second window sizes segmentation (n = 1,688). In this design, three risk classes, including "Nominal Risk", "Increased Risk", and "High Risk", have been considered as the output or labels for the classification models. In the specific case of KNN and in order to select the best value of k (i.e., the number of data points nearest to each data point used for majority voting) a comparison between twenty-seven outcomes of K values was conducted. As a result, K = 1 could provide the most accurate classification outcome as shown in Figure 2.

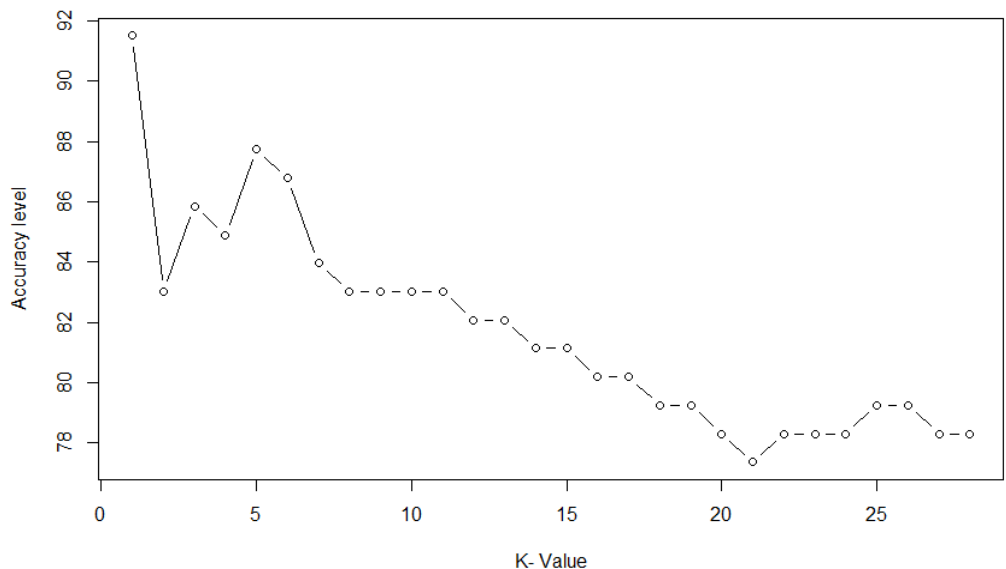

**Figure 2.** Selection of appropriate k value for KNN classification.

The sEMG signal data were split into training (75%) and testing sets (25%). The testing dataset was used to calculate the accuracy rate of correctly classified data points that were kept out during training. A 10-fold cross-validation was used in this study to provide validation outcomes of 10 replications of the training and testing datasets. Specifically, 10 combinations of each dataset were run and replaced iteratively with a low variation of 2% between 10 runs. After 10 iterations of both datasets, the average classification accuracies were recorded in Table 4.

**Table 4.** Classification testing accuracy (%) of machine learning models.

| Time Segmentation | Decision Tree | SVM | KNN | Random Forest |
|---|---|---|---|---|
| 1-sec (n = 425) | 97.70 | 97.17 | 91.51 | 96.23 |
| 0.5-sec (n = 845) | 99.35 | 98.90 | 91.87 | 99.24 |
| 0.25-sec (n = 1,688) | 99.05 | 99.00 | 96.21 | 99.21 |

Considering the results in Table 4, the Decision Tree algorithms manifested the best performance for classifying lifting activities based on the risk assessment using the NIOSH equations across two time-segmentations: 1-second (i.e., 97.70% accuracy) and 0.5-second (i.e., 99.35% accuracy). Table 5 provides the confusion matrix of the classification results using the Decision Tree across three time-segmentations.

**Table 5.** Confusion matrix of classification results from decision tree.

| Risk Class | 1-Sec Segmentation | | | 0.5-Sec Segmentation | | | 0.25-Sec Segmentation | | |
|---|---|---|---|---|---|---|---|---|---|
| | NR | IR | HR | NR | IR | HR | NR | IR | HR |
| Nominal Risk (NR) | 34 | 0 | 0 | 68 | 0 | 0 | 137 | 0 | 0 |
| Increased Risk (IR) | 0 | 19 | 3 | 0 | 42 | 1 | 0 | 77 | 4 |
| High Risk (HR) | 0 | 0 | 50 | 0 | 0 | 100 | 0 | 0 | 204 |

In the 0.25-sec segmentation, Random Forest provided the most accurate classification result (99.21%) compared to the other three methods: Decision Tree (99.05%), SVM (99%), and KNN (96.21%). The "Increased Risk" class had the highest misclassification rate compared to the "Nominal Risk" and "High Risk" classes in all three window size segmentation cases.

## 5. Discussion

Results indicate that the level of physical stress and WMSD risks represented by the NIOSH Lifting Index can be detected using machine learning models trained with sEMG sensor data. LI value below 1.2 as characterized in the first 19 experiments correspond to an average peak value of around 150 µV and 165 µV (time domain) and 20 µV and 21 µV (frequency domain). The higher values of weight and correspondingly LI are associated with higher values of average peak with the highest being 250µV (time domain) and 33.56µV (frequency domain). Therefore, this correlation can be leveraged to determine the risk level of lifting task according to the NIOSH lifting equation and using machine learning models. This relationship is depicted in the graph shown in Figure 3 where the sEMG values of the TH are plotted against the LI and risk levels are color-coded.

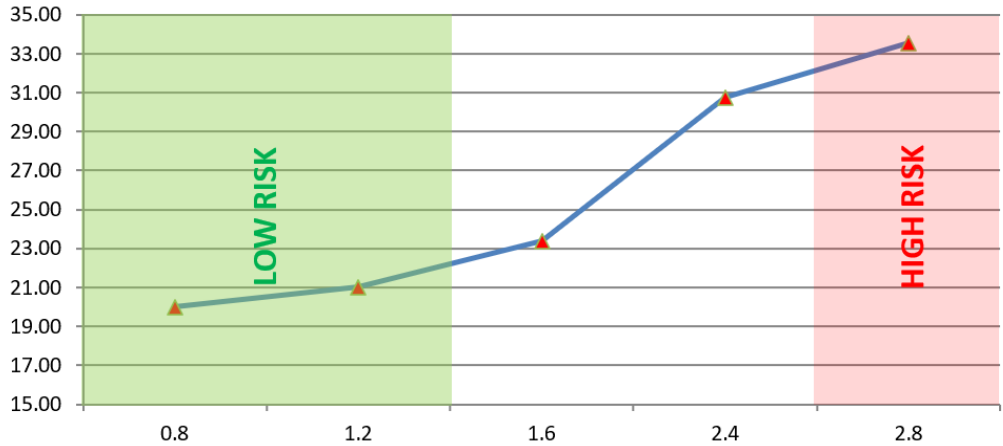

**Figure 3.** Fourier transformed averages of peak values (µV) on horizontal axis versus the LI on vertical axis.

The results of machine learning classification show that the decision tree model tends to provide the most accurate classification results of the collected EMG signal data using the NIOSH lifting equation. In addition, the "Increased Risk" class has the highest misclassification error with the "High Risk" class. While the classification testing accuracy at 0.5-seconds segment size reaches its highest value, the 0.25- and 1-seconds ones show relatively comparable results. The choice of the segmentation size dictates the level of information packed in one window but also affects the number of data points used to train machine learning models. Across three time-segmentations, the results using KNN model showed the highest misclassification error compared the other three models (i.e., lowest overall accuracy rate). The testing accuracies of KNN model were recorded from 91.51% (in 1-second segmentation) to 96.21% (in 0.25-second segmentation). To rule out the possibility of overfitting in the case of KNN with K=1, another test with K=5 (which showed the second highest accuracy in Figure 2) was conducted. According to Table 6, KNN with K=5 accuracies fail to outperform those of the model with K=1, so it is highly unlikely that overfitting has been an issue in case.

**Table 6.** Classification testing accuracy (%) for KNN with K = 5.

| Time Segmentation | KNN |
| --- | --- |
| 1-sec (n = 425) | 87.79 |
| 0.5-sec (n = 845) | 92.89 |
| 0.25-sec (n = 1,688) | 93.60 |

## 6. Conclusions

Appropriate body posture in heavy work and particularly during material handling activities can prevent the risk of WMSDs, minimize days away from work, and consequently improve the job productivity and safety records. In this study, the surface electromyogram (sEMG) sensors are employed to enable the automatic detection of harmful lifting activities using NIOSH lifting equation. A total number of 54 simple lifting tests have been conducted using adjustable weight in six categories with various three risk levels defined based on the NIOSH lifting equation. Results show that the level of risk characterized by the the NIOSH Lifting Index and measured by the pressure on the upper back (i.e., Thoracic or TH) muscles using the sEMG sensors can be determined using machine learning models. Towards this goal, four machine learning classification models (i.e., SVM, KNN, Decision Tree, and Random Forest) have been trained using the sEMG data points and risk levels. Decision Tree algorithm resulted in the highest test accuracy level of 97.70%, 99.35%, and 99.05% across three time-segmentations of 1-second, 0.5-second, and 0.25-second, respectively. The level of physical stress and WMSD risks associated with lifting tasks are represented by the NIOSH Lifting Index. LI value below 1.0 denotes a Normal Risk to a healthy employee. As the LI increases, the level of associated risk increases correspondingly and once the LI value exceeds 3.0 it represents a higher risk associated with the task. sEMG amplitude difference related to a single muscle signifies the varying level of muscle activities.

This study contributes to the body of knowledge and practice in industries such as construction where lifting is a major part of manual material handling. It investigates ergonomic postural hazards using sEMG sensors and enables classification of risk based on the lifting equation developed by NIOSH using four different machine learning methods. The classification results confirmed that there are distinguishable patterns in the sEMG signal data based on the weight in a lifting activity, which proves useful to provide insights regarding recognition of ergonomically hazardous body postures. This study, however, is limited since the experiments were conducted by only one person. Future work should increase the number of participants from which data is collected and conduct subject-dependent and subject-independent tests to evaluate generalizability of the results. Models that are more generic in their function can then be obtained to develop applications that are ready to deploy in the field. Real-world experiments in uncontrolled environments can also contribute to this goal.

**Conflicts of Interest:** The authors declare no conflict of interest.